\documentclass[conference]{IEEEtran}
\IEEEoverridecommandlockouts
\usepackage{cite}
\usepackage{amsmath,amssymb,amsfonts}
\usepackage{algorithmic}
\usepackage{graphicx}
\usepackage{textcomp}
\usepackage{xcolor}
\usepackage{subcaption}
\usepackage{booktabs}
\usepackage{algorithm}
\def\BibTeX{{\rm B\kern-.05em{\sc i\kern-.025em b}\kern-.08em
    T\kern-.1667em\lower.7ex\hbox{E}\kern-.125emX}}

\usepackage{multirow}
\usepackage{url}

\begin{document}

\title{Feature Attribution Explanations for Spiking Neural Networks}

\author{\IEEEauthorblockN{Elisa Nguyen}
\IEEEauthorblockA{\textit{University of Twente}\\
Enschede, Netherlands \\
0000-0003-0224-268X}
\and
\IEEEauthorblockN{Meike Nauta}
\IEEEauthorblockA{\textit{University of Twente} \\
Enschede, Netherlands \\
0000-0002-0558-3810}
\and
\IEEEauthorblockN{Gwenn Englebienne}
\IEEEauthorblockA{\textit{University of Twente} \\
Enschede, Netherlands \\
0000-0002-3130-2082}
\and
\IEEEauthorblockN{Christin Seifert}
\IEEEauthorblockA{\textit{University of Marburg} \\
Marburg, Germany \\
0000-0002-6776-3868}
}

\maketitle

\begin{abstract}
Third-generation artificial neural networks, Spiking Neural Networks (SNNs), can be efficiently implemented on hardware. Their implementation on neuromorphic chips opens a broad range of applications, such as machine learning-based autonomous control and intelligent biomedical devices. In critical applications, however, insight into the reasoning of SNNs is important, thus SNNs need to be equipped with the ability to explain how decisions are reached. We present \textit{Temporal Spike Attribution} (TSA), a local explanation method for SNNs. To compute the explanation, we aggregate all information available in model-internal variables: spike times and model weights. We evaluate TSA on artificial and real-world time series data and measure explanation quality w.r.t. multiple quantitative criteria. We find that TSA correctly identifies a small subset of input features relevant to the decision (i.e., is output-complete and compact) and generates similar explanations for similar inputs (i.e., is continuous). Further, our experiments show that incorporating the notion of \emph{absent} spikes improves explanation quality. Our work can serve as a starting point for explainable SNNs, with future implementations on hardware yielding not only predictions but also explanations in a broad range of application scenarios. Source code is available at \url{https://github.com/ElisaNguyen/tsa-explanations}.
\end{abstract}


\section{Introduction}
Spiking neural networks (SNNs), also known as third-generation artificial neural networks~\cite{Maass.1997}, consist of spiking neurons.
Spiking neurons emit spikes at certain points in time to transmit information, similar to action potentials in biological neurons and are thus close to biological reality~\cite{Gerstner.2014}. SNNs are at least as powerful as deep artificial neural networks with continuous activation functions (ANNs)~\cite{Maass.1997}. Their applicability to supervised, unsupervised and reinforcement learning are active research areas~\cite{Ponulak2011-tt}. However, the predictive performance of SNNs is not yet on par with ANNs due to the non-differentiability of spikes, making SNN optimization an active research field~\cite{Wang.2020}. Nonetheless, SNNs are interesting as they yield the potential to be implemented in neuromorphic hardware, which is energy- and memory-efficient~\cite{Murray.1998}.
    Moreover, studies show improved adversarial robustness of SNNs~\cite{sharmin.2019}. Their inherent temporal nature also lends itself naturally to processing temporal data making them suitable for critical domains relying on sensor data such as autonomous control~\cite{Bing.2018} and applications using biomedical signals~\cite{azghadi.2020}.
    
    Critical domains, for example medical applications, pose specific requirements to machine learning models. In addition to high predictive performance, models should make predictions based on the right reasons and be transparent about their decision-making process~\cite{He.2019}. Exposing important information of machine learning models is the focus of research on EXplainable Artificial Intelligence (XAI)~\cite{Adadi.2018, Guidotti.2019}. Model explanations address the requirement for algorithm transparency and provide methods to inspect model behavior~\cite{Molnar.19102020}. 
    While various explanation methods exist for second-generation artificial neural networks (ANNs)~\cite{Guidotti.2019, Adadi.2018}, to the best of our knowledge, the current body of work in explaining SNNs only comprises two major works, namely \cite{Jeyasothy.28022019,kim2021}. 
    XAI for SNNS is thus yet a sparsely studied research area. 
    If left unaddressed, this research gap could lead to situations where SNNs are methodologically mature for real-world deployment but remain unused because they lack transparency.

   We contribute to the field of XAI for SNNs by presenting Temporal Spike Attributions (TSA), an SNN-specific explanation method. The resulting explanations are \emph{local}, i.e., explain a particular prediction and answer the question: `Why did the model make this decision?'~\cite{Adadi.2018}. 
   We build on the explanation method of~\cite{kim2021} which uses the model's spike trains. We additionally include the SNN's weights to consider \emph{all} model-internal variables and regard the \emph{absence of spikes} to be informative as it also impacts the resulting spike patterns. 
   TSA results in more complete and correct explanations due to the utilization of comprehensive model-internal information. We demonstrate TSA on time series data. In contrast to anecdotal evidence which is mainly used to evaluate XAI methods~\cite{nauta2022anecdotal}, we systematically evaluate TSA  quantitatively w.r.t. multiple aspects relevant to explanation quality: The correctness of the explanations (correctness), the explanation's ability to capture the complete model behavior (output-completeness), sensitivity to input changes (continuity), and explanation size (compactness). 
   In summary, our contributions are as follows:
    \begin{enumerate}
        \item We present Temporal Spike Attribution (TSA), a local feature attribution method for SNNs inferred from all model internal variables.
        \item We apply Kim \& Panda's~\cite{kim2021} explanation method, which uses only spike train information, on time series data as a baseline to show the impact of incorporating all model internal information in TSA. 
        \item We thoroughly validate TSA's explanation performance using a multi-faceted quantitative evaluation of feature attribution explanations for SNNs, evaluating correctness, completeness, continuity and compactness.
    \end{enumerate}
    
    Because SNNs are more popular in neuroscience than machine learning, we briefly introduce SNNs in Section~\ref{sec:bg:snns}. Section~\ref{sec:rw} reviews related works on SNN explainability. We reflect on the effect of SNN-internal variables on a prediction and present our explanation method, TSA, capturing these effects in Section~\ref{sec:method}. The multi-faceted evaluation in Sections~\ref{sec:experiments} and~\ref{sec:results} shows the improved explanation performance of TSA. We discuss results in Section~\ref{sec:discussion} and conclude in Section~\ref{sec:conclusion}. 

\section{Spiking Neural Networks}
    \label{sec:bg:snns}

\begin{figure}[t!b]
        \centering
        \includegraphics[width=\linewidth]{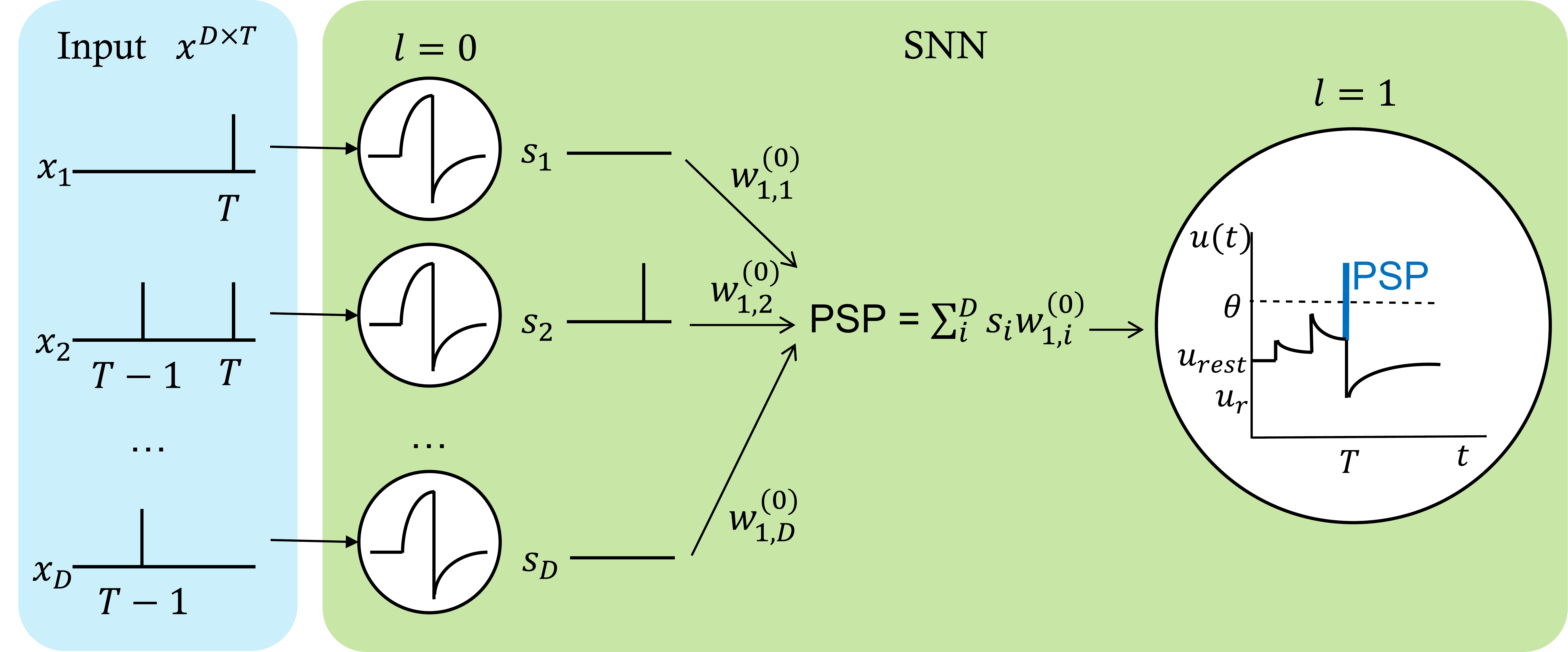}
        \caption{Schematic overview of an SNN with LIF neurons. Input spikes $x_i$ are fed to the SNN. Internally, it transmits information as spike patterns $s_i$, which are propagated forward with weights $w_{j,i}^{(l)}$ to determine the postsynaptic potential (PSP). The state of the postsynaptic neuron $u_j$ is changed by the PSP.}
        \label{fig:snn_overview}
    \end{figure}

    This section introduces SNNs and their components. 
    SNNs are characterized by their computational units, spiking neurons~\cite{Gerstner.2014}. Analogous to traditional ANNs, SNNs are networks of spiking neurons with weighted connections.
    SNNs process information in the form of \emph{spike trains}, i.e., spikes over time. Neuron models differ in the spike generation mechanisms. Our proposed method is independent of the chosen spike generation mechanism. Without loss of generality, we employ the commonly used leaky integrate-and-fire (LIF) neuron model (cf. Figure~\ref{fig:snn_overview}). The membrane potential $u$ of a LIF neuron can be modeled with a linear differential equation:
    
      \begin{equation}
      \label{eq:lifintegrate}
        \tau_{m}\frac{d u}{d t} = -[u(t)-u_{\textrm{rest}}] + I(t),
    \end{equation}
    
    where $I(t)$ describes the amount by which $u$ changes to external input. $\tau_m$ is the time constant of the neuron, which dictates the decay of $u$ in time. 
    A LIF neuron \emph{fires} when $u(t)$ crosses a threshold $\theta$ from below. Upon firing a spike, the membrane potentials of the postsynaptic neurons are changed by the weight value, as the spikes are propagated forward in the network. The sign of the weight defines the synapses' nature (i.e., inhibitory or excitatory) and the weight value defines the strength of the postsynaptic potential. 
    After firing, $u$ is reset to a low reset potential $u_{r}$, and then slowly increases back to its resting value $u_{\textrm{rest}}$. 
    
    SNNs internally process spike trains and thus require input data in the form of spikes. The translation of non-spiking to spiking data is called \emph{neural coding}. Different neural codes exist with temporal and rate coding being the most common. 
    Temporal coding is biologically more plausible than rate coding because it emphasizes exact spike times as information carriers~\cite{Gerstner.2014}. 

\section{Related Work}
\label{sec:rw}
        
Our work is positioned in the field of EXplainable Artificial Intelligence (XAI) for SNNs. 
XAI researches methods to address the black-box nature of machine learning models and explain their reasoning to laypersons and experts ~\cite{Molnar.19102020}. 
Machine learning models can either be explained globally by providing an overview of the whole model, or locally by explaining single predictions. Global explanations aim at providing a global understanding of how input relates to an outcome distribution addressing the model explanation and model inspection problem~\cite{Adadi.2018, Guidotti.2019}. Examples of global explanation methods are \cite{pmlr-v80-kim18d} and \cite{Nauta2021NeuralPT}.
The complexity of global explanations increases with the number of input features and model parameters and is therefore a challenging problem in XAI. 
Local explanations target individual model predictions and address the outcome explanation problem~\cite{Adadi.2018, Guidotti.2019}, i.e. they explain the relation between a specific model input and output. Two prominent examples of local explanations are LIME~\cite{ribeiro_2016_lime} and SHAP~\cite{lundberg_lee_2017_shap}. 
We aim to explain predictions of SNNs and develop a local explanation method.

\subsection{Explaining Spiking Neural Networks}    

\begin{table}[t]
    \centering
    \caption{Comparison of our method (TSA) to related work FSF~\cite{Jeyasothy.28022019} and SAM~\cite{kim2021} in terms of XAI taxonomy~\cite{Adadi.2018} and evaluation~\cite{nauta2022anecdotal}. Completeness refers to output-completeness.}
      \begin{tabular}{@{}lp{0.18\linewidth}p{0.2\linewidth}p{0.3\linewidth}@{}}
        \toprule
         & {FSF}~\cite{Jeyasothy.28022019} & {SAM}~\cite{kim2021} & {TSA (Ours)}\\
        \midrule
        \textbf{XAI Method} & Post-hoc & Post-hoc & Post-hoc \\
        \textbf{Scope} & Global & Local& Local\\
        \textbf{Data type} & Tabular & Images & Time-series \\
        \textbf{Model}  & MC-SEFRON & Convolutional SNNs & Specific to SNNs \\
        \textbf{Evaluation} \\
        Correctness & $\checkmark$ & $\times$ & $\checkmark$\\
        Completeness & $\times$ & $\times$ & $\checkmark$\\
        Continuity & $\times$ & $\times$ & $\checkmark$\\
        Compactness & $\times$ & $\times$ & $\checkmark$\\
        Coherence & $\times$ & $\checkmark$ & $\times$\\
        \bottomrule
    \end{tabular}
    \label{tab:comparison_rw}
    \vspace{-1em}
    \end{table}  
    
        Few works have studied explaining SNNs, which we introduce in the following.
    Jeyasothy et al.~\cite{Jeyasothy.28022019} present feature strength functions (FSFs) to explain a specific SNN model architecture MC-SEFRON with a population encoding layer, no hidden layers, and time-dependent weights. FSFs invert the population coding scheme to link the explanation back to input features and extract interpretable knowledge. 
    FSFs are functions of the input, i.e. in a human-understandable domain rather than the temporal domain of spike trains. The FSFs are a global and model-specific explanation method, which addresses the model inspection problem~\cite{Guidotti.2019}. In contrast, we target local explanations to explain model decisions that are applicable to a wider range of SNN models. 

    Kim \& Panda~\cite{kim2021} present \textit{Spike Activation Map} (SAM), a local explanation method for SNNs. 
    SAM generates visual heatmaps based on a calculation of input feature importance (in the image classification case, these are pixels) and was studied on deep convolutional SNNs with LIF neurons on image data.
    SAM is inspired by the biological observation that short inter-spike intervals are deemed to carry information because they likely cause a postsynaptic spike. The authors define contribution scores of single spikes and aggregate them to represent the spike train contribution in the neuronal contribution score (NCS). 
    The final activation map is computed at time $t$ by a forward pass in the network by multiplying  NCS's at $t$ and summarizing NCS's across the channel axis of convolutional layers.
    In contrast to the model-specific FSFs~\cite{Jeyasothy.28022019}, NCSs are model-agnostic because they are solely based on spike information which is part of all SNNs.  
    
    Our explanation method TSA is model-agnostic but not model-independent because TSA also takes the model's weights into account, which represent what the SNN has learned. 
    Furthermore, we aim to cater to the intrinsic temporal design of SNNs and therefore designed TSA for explaining predictions of a time series classification task. We look at time series data as opposed to image data to better fit the temporal nature of SNNs. Thus, we contribute to local explanations for SNNs and compare TSA to SAM. 
    
    Table~\ref{tab:comparison_rw} presents a concise comparison of the related work and our explanation method. We do not compare to model-agnostic methods for ANNs (e.g., LIME~\cite{ribeiro_2016_lime}) because their application to SNNs is not trivial. Moreover, ANN-based explanations do not rely on SNN model internals and hence might not capture the true model behavior~\cite{Poyiadzi2021OnTO}. Our aim is
    an SNN-specific explanation method.

\subsection{Evaluating SNN Explanations}
    In contrast to evaluating the predictive performance of models with quasi-standard evaluation metrics (e.g., F-score, AUC), evaluating explanations is an ongoing research topic.
    Since the recipients of explanations are humans and are usually context-dependent, there is no standard evaluation protocol~\cite{Molnar.19102020}.
    In addition, a good explanation fulfills several different properties, e.g. the correctness or faithfulness in explaining the model behavior and human-comprehensibility among others~\cite{nauta2022anecdotal,ijcai/BhattWM20,ijcai/0002C20}. 
    Moreover, evaluating explanations is challenging because the ground truth (what the model actually learned) is rarely known. To overcome this issue, one could apply the ``Controlled Synthetic Data Check''~\cite{nauta2022anecdotal} where a model is applied to (structured) synthetic datasets such that the true data distribution is known, e.g.~\cite{liu_synthetic_2021}. We apply this method by constructing a synthetic data set for a classification problem on two input sensors (cf. Section~\ref{ssec:experiments:datasets}). 

    The evaluations of FSFs and SAM were each focused on one aspect of explanation quality: Jeyasothy et al.~\cite{Jeyasothy.28022019} evaluated ``reliability" by using FSFs instead of model weights in the same prediction task, testing how correctly the FSFs capture the global model behavior. Kim \& Panda~\cite{kim2021} tested the ``accuracy" of SAM by comparison with an existing heatmap explanation on ANNs, i.e. how coherent and aligned SAM explanations are to other explanations. 
    In our work, we perform a multi-faceted evaluation based on the Co-12 framework for evaluating XAI~\cite{nauta2022anecdotal} on a synthetic and a real-world data set. More specifically, we evaluate correctness, output-completeness, continuity, and compactness as defined by Nauta et al.~\cite{nauta2022anecdotal}. We chose this subset of Co-12 properties as we focus on studying the content of TSA explanations first before considering presentation- and user-related properties.

\section{Temporal Spike Attribution (TSA)}
    \label{sec:method}
    SNNs learn internal weights during training and process data as spike trains (cf. Section~\ref{sec:bg:snns}.
    Whereas the Spike Activation Map (SAM)~\cite{kim2021}
    considers only spike trains to generate explanations, TSA captures \emph{all} information available in the model for a prediction $\hat{y}$ at time $T$ of one $D$-dimensional input $x^{D\times T}$. These information are (i) spike times $S^{(l)}$, (ii) learned weights $W$, and (iii) membrane potentials at the output layer $U^{(L)}$. Each component has an influence on the output $\hat{y}$ and therefore should be included in the explanation.
    We explain the single components in Sections~\ref{subsec:spiketimes} to~\ref{subsec:outputpotential} and describe their integration to a feature attribution explanation in Section~\ref{sec:attribution_formula}.

    \subsection{Influence of Hidden Neuron Spike Times}
        \label{subsec:spiketimes}
        In temporal coding, the information about the data is assumed to be in the exact spike times of a neuron~\cite{Gerstner.2014}. The spike times indicate the attribution of neurons $I$ to their downstream neurons $J$ and represent the model's activation in a prediction. Hence, the spike times influence the prediction.
        The relationship between the spike times of $I$ and their attribution to $J$ is captured in~\cite{kim2021}'s neuronal contribution score (NCS). The NCS is characterized by $\gamma$, which specifies the steepness of the exponential decay over time. We define the decay at the same rate as the decay of the LIF neuron's membrane potential $u$ to reflect the dynamics of the model.
        
        While we build on the NCS, our \textit{spike time component} $N_{i}(t)$ additionally considers the absence of spikes as information carriers. In a fully connected SNN, each neuron of layer $l$ is connected with each neuron of layer $l+1$. The weighted sum of the neuron's spiking behavior in $l$ determines the amount by which the membrane potentials of neurons in $l+1$ are changed. Absent spikes do not contribute to this sum. Hence, if a neuron $i$ does not spike at time $t$, it does not contribute actively to a change in $J$, allowing a natural decay. Absent spikes can thus be understood in two ways: (i) an absent spike does not affect $u_j$ (cf. Eq.~\ref{eq:ingredient_spikes_s}) or (ii) an absent spike affects $u_j$ by \emph{not} changing $u_j$ and letting it decay naturally (cf. Eq.~\ref{eq:ingredient_spikes_ns}). 
        In the second case, the attribution of absent spikes to the postsynaptic neuron is negative. However,
        absent spike attribution should not weigh as much as spikes because their effect is highly dependent on other incoming synapses. We weigh the contribution of absent spikes by $\frac{1}{B}$ as an approximation of their attribution factor, with $B$ being the size of the preceding layer. This approximation is simple and reflects the relative magnitude of a non-spiking neuron's attribution. Formally, we calculate the spike time component $N_{i}(t)$ as follows:    
          \begin{align}
          \small
            \label{eq:ingredient_spikes_s}
            N^{S}_{i}(t) &= \sum_{t'=0}^{t}
            \begin{cases}
                \exp (-\gamma |t - t'|) & \text{if } x_{i, t'}=1\\
                0 & \text{Otherwise}
            \end{cases} 
            \end{align}
        only considering the presence of spikes, and 
            \begin{align}
            \small
            \label{eq:ingredient_spikes_ns}
            N^{NS}_{i}(t) &= \sum_{t'=0}^{t}
            \begin{cases}
                \exp (-\gamma |t - t'|) & \text{if } x_{i, t'}=1\\
                -\frac{1}{B}\exp (-\gamma |t - t'|) & \text{Otherwise}
            \end{cases}
            \end{align}
            when including the information about absent spikes.

        In our experiments, we compare both
        variants as TSA\textsuperscript{S} (spikes only according to Eq.~\ref{eq:ingredient_spikes_s}), and TSA\textsuperscript{NS} (non spikes included, according to Eq.~\ref{eq:ingredient_spikes_ns}).

\subsection{Influence of Model Weights}
        \label{subsec:weights}
        The weights $W$ represent the strengths of connections of an SNN and have so far not been considered in feature attribution explanations for SNNs. We specifically include weights to capture the contribution of connections to the predictions of an SNN. 
        Weights determine the impact of spikes to downstream neurons $J$ directly, where the weight value indicates the weight's attribution to a neuron $j \in J$, and the sign specifies whether the synapse is excitatory or inhibitory and leads to an in- or decrease of $u_j$. 
        Since weights are a property of the model, i.e. independent of the input, the weight contribution obtains its meaning in combination with other components. 
            
    \subsection{Influence of Output Layer}
        \label{subsec:outputpotential}
        The output layer is the last computational layer, i.e., the basis for prediction. The output layer consists of spiking neurons, thus the prediction is dictated either by spike patterns $S^{(L)}$ or membrane potentials $U^{(L)}$. In the first case, the influence of the output layer can be captured by the NCS. 
        The SNN models in our work follow \cite{Neftci.2019}'s architecture and make predictions based on the latter, i.e. the largest $u_i^{(L)}$ determines the predicted class. We capture the effect of the output layer on a model prediction by considering the classification softmax probability $P(t)$ in the computation of an explanation.
        \begin{equation}
            \Vec{P}(t) = \text{softmax}(U^{(L)})
        \end{equation}
    
    \subsection{Computation of TSA}
    
    \label{sec:attribution_formula}
        TSA combines neuron spike times, model weights, and output layer information in a forward pass into a final score as shown in Algorithm~\ref{alg:tsa}.
        A neuron $i$ generates spike train $s_i$ to the downstream computational layers. The neuron is connected to the next layer $l+1$ via the weight matrix $W^{(l)}$.
        Spike time information is captured by $\Vec{N}^{(l)}(t)$, model weights are contained in $W$, and output layer information, i.e. membrane potentials are encoded in $\Vec{P}(t)$.
        The spike times and the weights are combined by multiplying the diagonal matrix of $\Vec{N}^{(l)}(t_c)$ with $W^{(l)}$ per layer (line 7 in Algorithm~\ref{alg:tsa}). 
        The resulting matrix can be computed for the input layer and all hidden layers. The result is a set of matrices consisting of scores
        that represent how the presynaptic neurons contribute to the postsynaptic neurons under direct consideration of the weights. 
        The values are aggregated in a forward pass through the model. This represents how the input influences the model's neurons and is captured by the input contribution $C_I(t)$ (Line 8).
        The final feature attribution $A(x, t) \in \mathbb{R}^{O\times D \times t}$ is computed by multiplying $C_I(t)$ with the softmax probabilities (Line 10). The absence of spikes can be understood as either not affecting (Eq.~\ref{eq:ingredient_spikes_s}) or affecting downstream neurons (Eq.~\ref{eq:ingredient_spikes_ns}). We compare both
        variants as TSA\textsuperscript{S} (spikes only), and TSA\textsuperscript{NS} (non spikes included).
        Thus, the computation in line 6 of Algorithm~\ref{alg:tsa} differs respectively.
        
        \begin{algorithm}[t]
            \caption{\textbf{Temporal Spike Attribution}}\label{alg:tsa}
             Let $x\in\mathbb{R}^{D\times T}$ be an input to SNN $f$ with $L$ layers, $S^{(l)}$ a layer's spike trains, $U^{(L)}$ the output layer's membrane potentials, $W^{(l)}$ the weight matrix connecting layers $l$ and $l+1$, and $t$ the explanation time.
            \begin{algorithmic}[1]
                \STATE $S^{(1)}, ..., S^{(L-1)}, U^{(L)}\gets f(x)$
                \STATE $\Vec{P}(t) \gets \text{softmax}(U^{(L)})$ 
                \FOR{$t' = 0, 1, 2, ..., t$}
                    \STATE Initialize $C_I(t') = I \in \mathbb{R}^{DxD}$
                    \FOR{$l = 1, 2, ..., L-1$}
                        \STATE Compute $\Vec{N}^{(l)}(t')$
                        \STATE $N_W^{(l)}(t') \gets \text{diag}[\Vec{N}^{(l)}(t')]\cdot W^{(l)}$ 
                        \STATE $C_I(t') \gets C_I(t') \cdot N_W^{(l)}(t')$     
                    \ENDFOR
                    \STATE $A(x, t') \gets C_I(t')\cdot \text{diag}(\Vec{P}(t'))$
                    \STATE Concatenate $A(x, t')$ to attribution map $A(x, t)$.
                \ENDFOR
            \end{algorithmic}
        \end{algorithm}

\section{Experimental Setup}
    \label{sec:experiments}

    We demonstrate TSA\textsuperscript{S} and TSA\textsuperscript{NS} on both synthetic and real-world data using fully connected SNNs of different depths. Additionally, we compare the quality of the extracted explanations to SAM~\cite{kim2021}. 
    
\subsection{Data Sets}
\label{ssec:experiments:datasets}

\begin{figure}[tb]
        \centering
        \begin{subfigure}{\linewidth}
        \small
        \centering
        \begin{tabular}{lll}
        \toprule
        Label  & $x_1$ & $x_2$ \\ 
        \midrule
        A & 0 & 0 \\
        B & 1 & 1 \\
        C & 0 & 1 \\
        D & 1 & 0 \\\bottomrule
        \end{tabular}
        \end{subfigure}  
         \begin{subfigure}[c]{0.7\linewidth}
        \includegraphics[width=\textwidth]{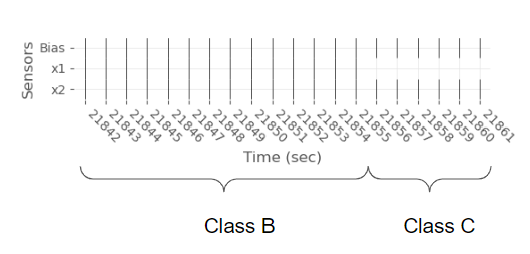}
        \end{subfigure}        
        \caption{Synthetic data set. Class label assignment (top) and example time series (bottom).}
        \label{fig:syn_data}
        \vspace{-1em}
     \end{figure}

    Synthetic data sets are commonly used in XAI research as the data's true distribution is known~\cite{liu_synthetic_2021}.
    For the \emph{synthetic} data set, we chose a simple task of classifying two-dimensional, binary time series data of varying length, i.e., $x_{i,t} \in \{0,1\}$, into one of four classes equivalent to a logical OR. An example is shown in Figure~\ref{fig:syn_data}. 
    The underlying idea of such a simple, synthetic dataset is that the SNN should learn and use the same reasoning as the data generation process, which is known a priori. We can then evaluate whether the explanation shows the expected reasoning. 
    We generate the data by sampling both the duration and activation of $x_i$ at random. The maximum duration for an activity is set at 600 time steps, and we generate 900,000 time steps sequentially as the entire data set. Once the data is generated, we add labels per time step according to the data. 
    We perform a 70\%-30\% sequential train-test split, i.e. the first 70\% of the data constitute the training and the remaining 30\% the test set. 
    
\begin{figure}[t]
        \centering
\includegraphics[width=0.95\linewidth,trim={0.9cm 1cm 1.5cm 3cm},clip]{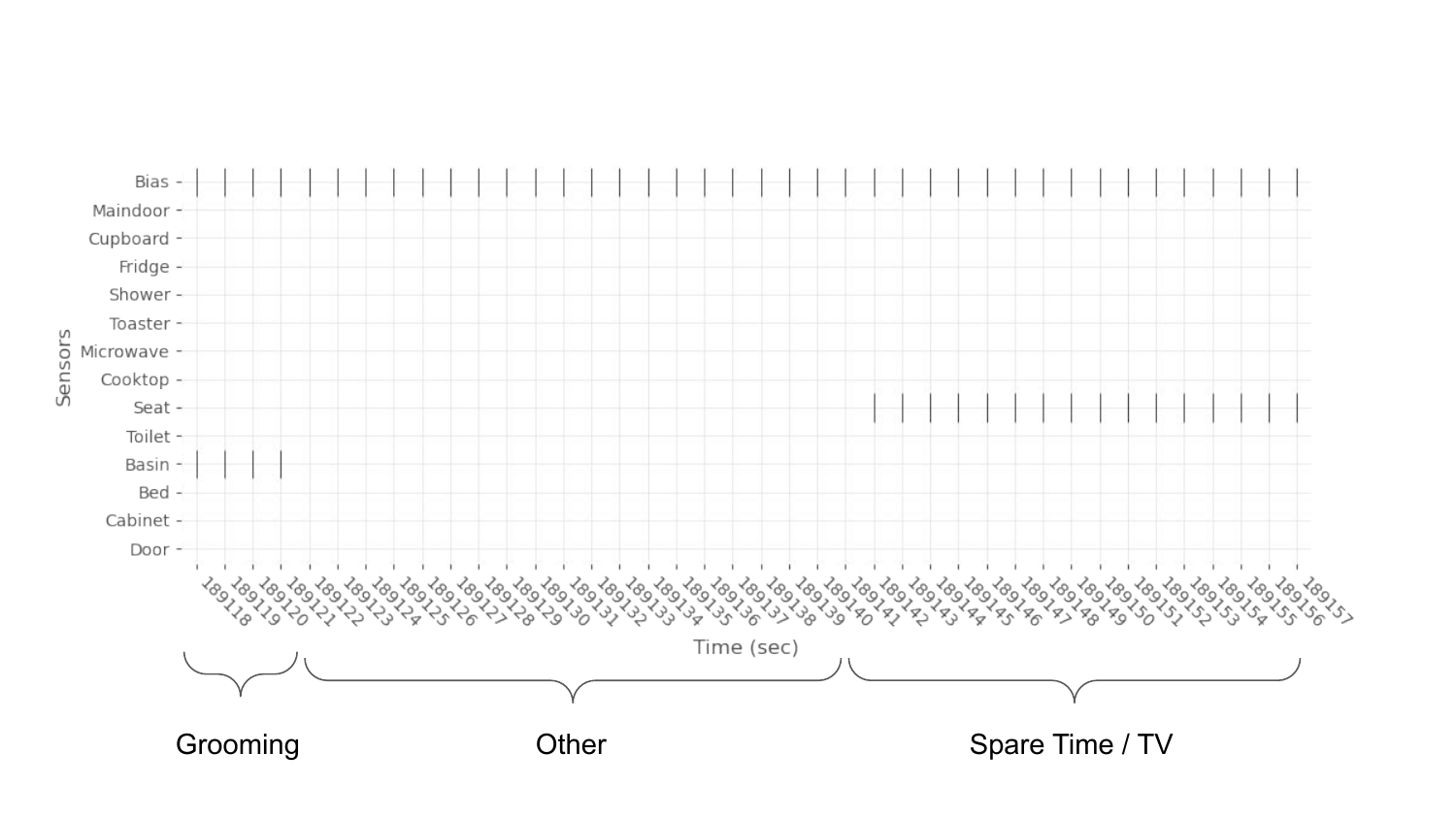}
        \caption{Data sample of subject A from the ADL dataset with either the Basin or the Seat sensor active.}
        \label{fig:adl_data_snippet}
        
        \end{figure}

    For a real-world scenario, we use the ``Activities of Daily Living Recognition using Binary Sensors'' data set (\emph{ADL})~\cite{Ordonez.2013}. ADL is an imbalanced multivariate time series data set that can be used for multi-class classification. The data was collected from a wireless binary sensor network installed in the homes of two subjects over 35 days at a time granularity of one second. The data set includes 10 classes specifying different activities of the subjects, e.g. \textit{Sleeping} or \textit{Toileting}. ADL is openly available in the UCI machine learning repository\footnotemark[1]. The sensors are human-interpretable, e.g., activation of the \textit{Bed} sensor means that the subject is lying in their bed (cf. Figure~\ref{fig:adl_data_snippet}). Since the data is human-understandable, feature attributions in this input space are easily understandable (e.g. ``The most attributing feature is the bed sensor activation at time $t$. Hence, it is important for the model that the subject lies in bed at $t$ for predicting the activity ``Sleeping'' at $t$.").
    The SNN models are trained to predict the activity at each time step. The input data is converted to spikes, and we apply no other data transformations. Thus, the neural coding is a direct mapping of spike times. As a preprocessing step, we add a constantly spiking sensor as a bias input to the data. 
    Gaps between activities were filled with inactivity in all sensors. We introduce the \textit{Other} class for these time gaps.\footnotemark[2]
    We treat the data as one long time series per subject and split the data set sequentially into a training (60\%), validation (20\%), and test (20\%) set to preserve temporal dependencies.

    \footnotetext[1]{https://archive.ics.uci.edu/ml/datasets/Activities+of+Daily+Living+\%28A\-DLs\%29+Recognition+Using+Binary+Sensors}

    \footnotetext[2]{We found two cases where the activity end precedes the start (Index 78, 80 of subject A). We excluded these activities from the data set as the data collection or labeling was faulty and filled their time with inactivity.}
    
\subsection{SNN Models}
    We train three SNNs with 1, 2 and 3 hidden layers denoted \textit{SNN-1L}, \textit{SNN-2L} and \textit{SNN-3L} to evaluate the effect of network depth on explanation quality, where the respective size of the hidden layers is determined by hyperparameter tuning.
    
    The models are implemented as recurrent networks with binary activations with discretized formulas of the model dynamics as in~\cite{Neftci.2019}. We train the models in a similar fashion, i.e. using backpropagation with surrogate gradients.
    We adapt the training procedure to our data set that exhibits a temporal order (i.e., activities follow one after another in time). The membrane potential state $u(t)$ is retained between data samples to reflect the temporal dependencies. So, the model state is initialized with the last state of the last simulation. 
    
    While the main focus of our work is explaining SNNs and not their optimization, the models should perform reasonably well, so that the models have learned something worth explaining. Therefore, we do a hyperparameter search in a greedy optimization process for 20 epochs under the assumption of independence of hyperparameters for the ADL task(Table~\ref{tab:tuning}). Due to the simplicity of the synthetic task, we omit hyperparameter tuning. 
    
        \begin{table}[t]
            \centering
            \small
            \caption{Results of greedy hyperparameter search. Tested ranges are: \{0.01, 0.001, 0.0001\} for $\Delta t$ and learning rate, \{0.1, 0.01, 0.001\} for $\tau_{syn}$ and $\tau_{mem}$, \{128, 256, 512\} for batch size, \{25, 50, 100, 200\} for hidden layer sizes.}
            \begin{small}
                \begin{tabular}{@{}lllll@{}}
                \toprule
                & \multicolumn{3}{c}{\textbf{ADL}} & \textbf{Synthetic}\\ 
                {Hyperparameter} & {SNN-1L} & {SNN-2L} & {SNN-3L} & {All SNNs}\\ \cmidrule(r){1-1}\cmidrule(r){2-4}\cmidrule(r){5-5}
                $\Delta t$ & 0.001 & 0.001 & 0.001 & 0.001\\
                $\tau_{syn}$                & 0.01           & 0.01         & 0.01    & 0.01  \\
                $\tau_{mem}$                & 0.01          & 0.001          & 0.01        & 0.001   \\
                Learning rate          & 0.01           & 0.001          & 0.001          &0.001   \\
                Batch size             & 128        & 256         & 512    &128       \\
                Hidden size 1 & -           &   100       & 50       &10   \\
                Hidden size 2 & -           & -           & 25        &10 \\ \bottomrule
                \end{tabular}
            \end{small}
            \label{tab:tuning}
            \vspace{-1em}
        \end{table}
    
    The final models were fully retrained on the training set with early stopping with a patience of 10 epochs, monitoring the validation loss.
    We compare the SNN models against the majority baseline, i.e., a classifier that always predicts the class most represented in the training set in terms of balanced accuracy~\cite{Mosley.2013} at the 95\% confidence interval (CI):
    \begin{equation}
        \small
        \textrm{CI} = 1.96\sqrt{\frac{\text{Balanced Accuracy}*(1-\text{Balanced Accuracy})}{N}}
    \end{equation}
    where  $N$ is the number of samples. The synthetic data set is balanced with four classes, thus the majority baseline has a balanced accuracy of .25, while in ADL the majority class (\textit{Sleeping}) leads to a balanced accuracy of .09. 
    
        \begin{table}[t]
          \centering
        \small
        \caption{SNN model performance for the synthetic and ADL task in balanced accuracy ()\% at 95\% CI. Results are based on one training run. The synthetic data set contains no validation split (n.a.). The baseline is a majority vote.}
          \begin{tabular}{clccc}
            \toprule
            Data & Model & Test & Train & Validation\\
            \midrule
           \multirow{4}{*}{\rotatebox{90}{\textbf{Synthetic}}}
            & SNN-1L & 98.3 $\pm$ 0.0 & 98.3 $\pm$ 0.0 & n.a. \\
            & SNN-2L & 95.6 $\pm$ 0.1 & 95.5 $\pm$ 0.1 & n.a. \\
            & SNN-3L &93.3 $\pm$ 0.1 & 93.2 $\pm$ 0.1& n.a. \\
            & Baseline & 25.0 & 25.0 & n.a. \\\midrule
           \multirow{4}{*}{\rotatebox{90}{\textbf{ADL}}}
            & SNN-1L & 51.6 $\pm$ 0.1 & 50.6 $\pm$ 0.1 & 53.6 $\pm$ 0.1 \\
            & SNN-2L & 51.7 $\pm$ 0.1 & 51.5 $\pm$ 0.1 & 54.9 $\pm$ 0.1 \\
            & SNN-3L & 50.0 $\pm$ 0.1& 49.0 $\pm$ 0.1& 52.0 $\pm$ 0.1  \\
            & Baseline & 9.1 &9.1 &9.1\\
          \bottomrule
        \end{tabular}
        \label{tab:model_perfomance}
        \vspace{-1em}
        \end{table}

Table \ref{tab:model_perfomance} reports model performance for both data sets. 
In the synthetic task, all models achieve high accuracy ($>$ 0.9), where model depth is correlated with lower performance. Since learning in SNNs is an active research area itself~\cite{Wang.2020}, the reason for this phenomenon is unclear. Still, all models are able to solve the synthetic task well and are therefore useful for our analysis of well-performing SNNs.
        All models learn the ADL task similarly well, significantly outperforming the baseline (selecting the majority class).
        while not overfitting (cf. Table~\ref{tab:model_perfomance}). While other work reports higher performance on ADL with complex ANNs~\cite{Hamad.2021}, our SNN models are sufficiently accurate for evaluating TSA in a real-world setting.

    \subsection{Evaluation}
        Because explanation quality is a multidimensional property (e.g., \cite{nauta2022anecdotal}), we quantify the explanation quality of TSA\textsuperscript{S} and TSA\textsuperscript{NS} objectively using mainly content-related properties of the Co-12 framework~\cite{nauta2022anecdotal}. 
        We use SAM as a baseline to investigate whether the incorporation of model weights improves explanations\footnotemark[3].
        To the best of our knowledge, we are the first to apply such quantitative evaluation measures to explanations for SNNs. 
        
        \footnotetext[3]{We note that \cite{kim2021} developed SAM to explain image data and did not claim that these maps are applicable to other data types. Still, we believe it is applicable as it uses only spike times, and applied it to our time series data sets.}
        
\subsubsection{Evaluation Setup}
\label{sec:eval_setup}
    
    Because we simulate our SNN models in our experiments as recurrent models with binary activations~\cite{Neftci.2019}, i.e. in a non-neuro\-morphic environment, calculating TSA on all test data is computationally not tractable.
    For an accurate and balanced evaluation of our explanations, we sub-sample the test data, choosing the same number of samples per class, to identify an explanation evaluation set. We assume that relevant features for the prediction are contained within a fixed time window of ${[t-1000, t]}$ where $t$ is the prediction time.
    For the synthetic data, we randomly select 25 samples per class (i.e., 100 overall). 
    For each ADL subject, nine samples across the start, middle, and end of an activity are chosen per class in the test set. 
    The start and end of the activity are defined as the first and last minute of this activity, respectively. Given these constraints, we sample $t$ at random, resulting in 180 samples (81 and 99 for subjects A and B).
    
        \subsubsection{Evaluation Measures}
        \label{subsec:eval_measures}
            \emph{Correctness} refers to whether the explanation reflects the true behavior of the model, which is universally desirable. 
            To measure correctness, we incrementally delete ranked feature segments (i.e., most attributing first) and record model prediction performance~\cite{nauta2022anecdotal}. The area under the curve of the resulting graph represents explanation correctness as \textit{explanation selectivity}~\cite{Montavon.2018}. A low score is desirable, as the model performance is expected to drop significantly when highly attributing feature segments are deleted. 
            We define feature segments as a number of continuous, strictly positively or negatively attributing time steps within one input dimension $d$ of $x$ that is at most 10 seconds long. This duration is assumed to capture the temporal dependencies at an appropriate information coarseness for both classification tasks posed by the synthetic and ADL data sets.
            Moreover, the attribution values are not expected to vary strongly within 10 seconds if they are either positive or negative. We implement this incremental feature deletion in temporal data as an inversion of the input spike train, similar to the perturbations proposed by \cite{Schlegel.2019} for the evaluation of XAI methods on time series data.
            
            \emph{Output-completeness} assesses to which extent the set of identified important features $F$ is sufficient for prediction $\hat{y}$, i.e., $F\rightarrow\hat{y}$. 
            A perfect output-complete explanation covers all important features relevant to the prediction but might include more features than necessary (cf. criterion compactness). 
            We evaluate output-completeness by deleting unimportant feature segments and reporting the model performance upon deletion~\cite{nauta2022anecdotal}. We define unimportant feature segments as having zero attribution because TSA produces unscaled feature attribution explanations for which an importance threshold is difficult to define. 
            Contrary to the evaluation of correctness, we do not invert the unimportant feature segments to delete them. For correctness, we incrementally change small parts of the data so that the perturbation is not large in the beginning but accumulates. For evaluating output-completeness, however, we delete many features at once. 
            If we inverted all the unimportant features, the data would become unrealistic and not reflect the original data distribution (e.g., multiple sensors active around the house while the subject cannot be in multiple locations at once) since we delete all unimportant features at once. 
            Instead, we shuffle the unimportant features randomly in the time domain to delete any effect the unimportant features have at their original time. This resembles the permutation importance method for structured data~\cite{breiman2001random} where the importance of features is measured by how much a score, in our case the model performance, changes upon random permutation of the features. If feature segments are truly unimportant, the model predictions should not change.
            
\emph{Continuity} refers to an explanation method's capability to generalize. An explanation method is continuous if it generates similar explanations for similar input. As non-continuous behavior is difficult for a user to understand, continuity is desirable.
We measure continuity by inspecting the stability for slight variations in the input data. We use \textit{max-sensitivity} defined as the maximum Frobenius norm of the difference between the explanations on original and slightly varied data~\cite{yeh.2019}:
\begin{small}
            \begin{equation}
                \label{eq: max-sensitivity}
                \textrm{Sens}_{\textrm{max}}(e, f, x, t, r) = \max_{||x'-x|| \leq r}||e(f,x', t)-e(f,x, t)||
            \end{equation}
        \end{small}
        where $e$ refers to the explanation, $f$ to the model, $x$ the input data, $t$ the current timestep and $r$ defines the neighborhood region in which perturbed data $x'$ is still considered as similar to $x$. 
We vary the data randomly by perturbing the duration of active sensors in the range of 10\% of the original duration. 
Such perturbations are realistic given the task of activity prediction where the task duration is not rigid and different instances of the same activity can have a different pace, e.g. taking a bit more time for the activity ``shower".
             
\emph{Compactness} refers to the size of an explanation, where a compact, in our case, sparse, attribution map is desirable. We measure the mean size of all extracted explanations as the sum of absolute attribution values for compactness: 
        \begin{equation}
            \textrm{Compactness}(A) = \frac{1}{N}\sum_i^D\sum_j^t |a_{i,j}|
        \end{equation}
        where $A$ is the matrix of dimension $D\times t$, i.e. input dimensionality $D \times$ the timestep to be explained. $N$ denotes the total number of explanations extracted for the experiment.

\section{Results}
\label{sec:results}
In addition to the quantitative analysis in Section~\ref{sec: results_quantitative}, we present visual examples of explanations in Section~\ref{sec:results_visual} to give an impression of TSA and analyze its coherence~\cite{nauta2022anecdotal}.
    
\subsection{Quantitative Results}
\label{sec: results_quantitative}

\begin{table*}[tp]
\centering
\small
\caption{Quantitative evaluation results at 95\% CI of our TSA\textsuperscript{S} and TSA\textsuperscript{NS} explanations compared to SAM~\cite{kim2021} on Synthetic and ADL. Arrows indicate the direction of better performance. Continuity measured as max-sensitivity (no CI).}
\begin{tabular}{clccccccc}
\toprule
     & Measure & TSA\textsuperscript{S}& TSA\textsuperscript{NS} & {Baseline (SAM)} &     & TSA\textsuperscript{S}& TSA\textsuperscript{NS} & {Baseline (SAM)}\\
 \cmidrule(r){1-5} \cmidrule(r){6-9}
\multirow{16}{*}{\rotatebox[origin=c]{90}{\textbf{Synthetic}}}
& \multicolumn{4}{@{}l}{\textbf{Correctness $\downarrow$}} & \multirow{16}{*}{\rotatebox[origin=c]{90}{\textbf{ADL}}} & \\
    & SNN-1L    & 0.404 $\pm$ 0.096 & \textbf{0.115 $\pm$ 0.063} & 0.628 $\pm$ 0.095 & & 0.086 $\pm$ 0.041 &    \textbf{0.006 $\pm$ 0.011}    &    0.329 $\pm$ 0.069 \\
    & SNN-2L    & 0.809 $\pm$ 0.077 & \textbf{0.426 $\pm$ 0.097} & 0.762 $\pm$ 0.083 & & 0.633 $\pm$  0.070  &   \textbf{0.170 $\pm$ 0.055}     &   0.655  $\pm$ 0.069 \\
    & SNN-3L    & 0.813 $\pm$ 0.076  & \textbf{0.520 $\pm$ 0.098} & 0.822 $\pm$ 0.075 & &  0.505 $\pm$ 0.073   &  \textbf{0.154 $\pm$ 0.053}       & 0.377 $\pm$ 0.071   \\\cmidrule(r){2-2} \cmidrule(r){3-5} \cmidrule(r){7-9}
    & \multicolumn{4}{@{}l}{\textbf{Output-completeness $\uparrow$}} \\
    & SNN-1L    & 0.726 $\pm$ 0.087 & \textbf{1.000 $\pm$ 0.000} & 0.737 $\pm$ 0.086 & &   0.996  $\pm$  0.009 &    \textbf{1.000 $\pm$ 0.000}    &     0.624 $\pm$ 0.071 \\
    & SNN-2L    & 0.692 $\pm$ 0.090  & \textbf{1.000 $\pm$ 0.000} & 0.724 $\pm$ 0.088 & & 0.658 $\pm$ 0.069 &    \textbf{0.957 $\pm$ 0.030}    &   0.649 $\pm$ 0.070 \\
    & SNN-3L    & 0.600 $\pm$ 0.096 & \textbf{0.989 $\pm$ 0.020} & 0.680 $\pm$ 0.091 & & 0.631 $\pm$ 0.070   &   \textbf{0.991 $\pm$ 0.014}    &          0.553 $\pm$ 0.073  \\\cmidrule(r){2-2} \cmidrule(r){3-5} \cmidrule(r){7-9}
    & \multicolumn{4}{@{}l}{\textbf{Continuity $\downarrow$}}      \\
    & SNN-1L    & \textbf{0.282} & 0.334 & 0.679 & &  0.652   &    \textbf{0.474}    &   2.715   \\
    & SNN-2L    & \textbf{0.021} & 0.023 & 0.863 & &  0.011    &   \textbf{0.009}    &   4.338 \\
    & SNN-3L    & \textbf{0.001} & 0.002 & 0.676 & &   \textbf{0.002}  &   \textbf{0.002}    &   53.287  \\\cmidrule(r){2-2} \cmidrule(r){3-5} \cmidrule(r){7-9}
    & \multicolumn{4}{@{}l}{\textbf{Compactness $\downarrow$}}     \\
    & SNN-1L    & \textbf{0.344 $\pm$ 0.009} & 0.411 $\pm$ 0.005 & 1.143 $\pm$ 0.029  & &  \textbf{ 0.730 $\pm$ 0.109 }   &  1.651 $\pm$   0.138   &    9.634 $\pm$ 5.337 \\
    & SNN-2L    & \textbf{0.007 $\pm$ 0.000} & 0.011 $\pm$ 0.000 & 0.760 $\pm$ 0.031 &  &   \textbf{0.002 $\pm$ 0.000}   &  0.003 $\pm$   0.000  &   2.183 $\pm$ 0.633    \\
    & SNN-3L    & \textbf{0.000 $\pm$ 0.000}\footnotemark[3] & 0.001 $\pm$ 0.000 & 0.321 $\pm$ 0.011 & &  \textbf{0.001 $\pm$ 0.000}    & \textbf{0.001 $\pm$ 0.000}  &   49.197 $\pm$ 293.630 \\\bottomrule
    \multicolumn{9}{@{}l}{$^\mathrm{3}$While very small, the explanation size is $>0$. This value is rounded.}
 \end{tabular}
\label{tab:quantitative_results}
\vspace{-1em}
\end{table*}

The results of the quantitative analysis are presented in Table~\ref{tab:quantitative_results}. Similar explanation performance trends can be observed from the synthetic and ADL experiments: 
In terms of correctness and output-completeness, TSA\textsuperscript{NS} is clearly superior to TSA\textsuperscript{S} and SAM. This shows that the SNN's output is also based on the absence of spikes and that considering the absence of spikes is relevant for the explanation (cf. detailed discussion in Section~\ref{sec:discussion}). TSA\textsuperscript{S}, however, does not show a clear improvement to SAM in correctness, implying that the additional consideration of weights to spike times is not significant for capturing the model's true behavior. 

SAM is slightly better than TSA\textsuperscript{S} in terms of output-completeness in the synthetic task, while TSA\textsuperscript{S} is better in the ADL experiments. Investigating the extent to which the inclusion of model weights in the computation of a feature attribution explanation for SNNs improves the explanation's completeness in finding all relevant parts of the model input is left for future work.
The results in the evaluation of continuity show that TSA\textsuperscript{S} and TSA\textsuperscript{NS} both are more stable than SAM in producing similar explanations for similar, but slightly different input data. Also in terms of explanation size, TSA\textsuperscript{S} and TSA\textsuperscript{NS} generate smaller explanations than SAM where TSA\textsuperscript{S} produces the most compact explanations. 

\subsection{Visual Inspection}
\label{sec:results_visual}
TSA generates feature attribution explanations, which can be visualized and overlaid with the input data. We visually inspect TSA explanations in the case of correct prediction, misclassification, and deep SNN models. In all examples, we display only the last 7 (synthetic) and 40 seconds (ADL) of the sample due to space constraints. The visualizations show positive and negative class attributions in red and blue, respectively. White corresponds to an attribution value of zero. The color scale is explanation-specific and dictated by the largest absolute attribution value. The $y$-axis shows the input dimensions, i.e. the sensors of the data set. Sensor activation is visualized by spikes (vertical lines) at a time step.
    
\subsubsection{Correct Predictions}

\begin{figure}[t]
            \centering
            \includegraphics[width=0.95\linewidth]{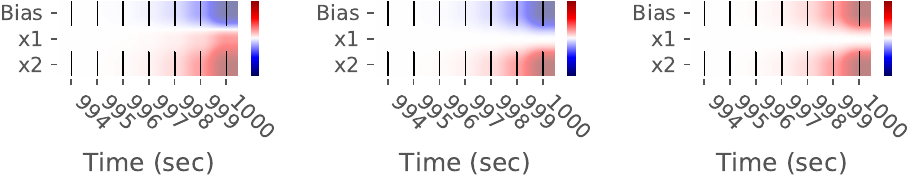}
            \caption{TSA\textsuperscript{NS} (left), TSA\textsuperscript{S} (center), and SAM (right) activation of class C for SNN-1L's correct prediction of $y=$C for the synthetic task. \textit{Best viewed in color.}}
            \label{fig:synthetic_viz}
            \vspace{-1em}
        \end{figure}

\begin{figure}[t]
            \centering
            \begin{subfigure}[b]{0.95\linewidth}
                \centering \includegraphics[width=\textwidth]{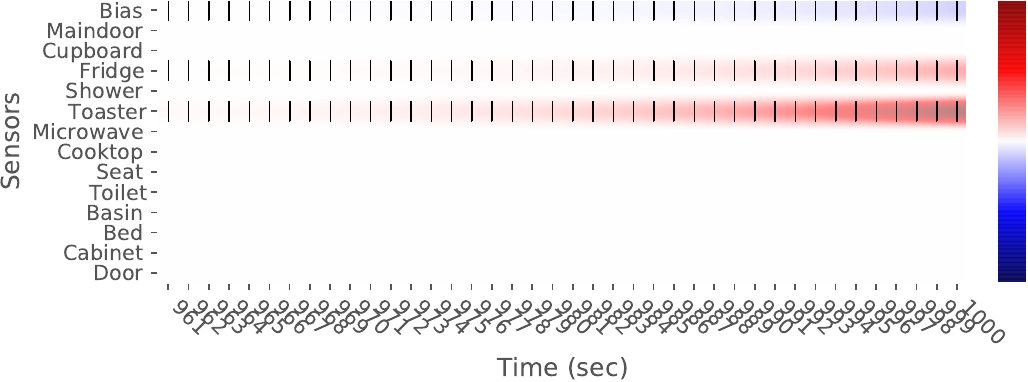}
                \caption{TSA\textsuperscript{S} breakfast class activation.}
            \label{fig:v_one_s}
            \end{subfigure}
            \hfill
            \begin{subfigure}[b]{0.95\linewidth}
                \centering \includegraphics[width=\textwidth]{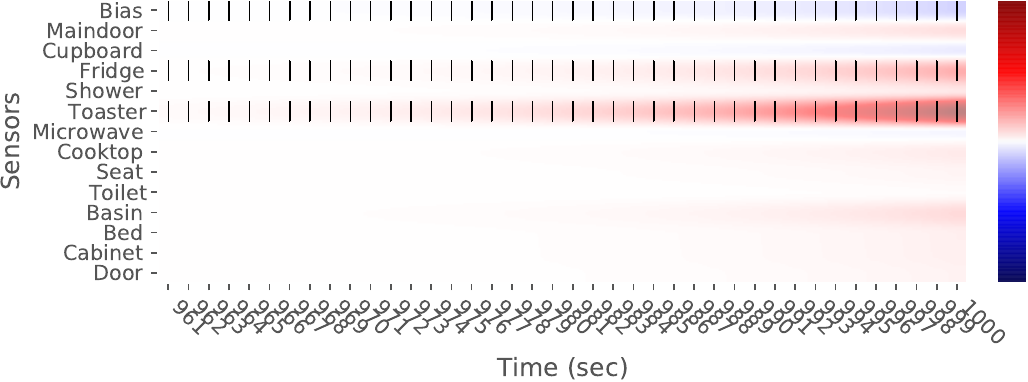}
                \caption{TSA\textsuperscript{NS} breakfast class activation.}
                \label{fig:v_one_ns}
            \end{subfigure}
            \hfill
            \begin{subfigure}[b]{0.95\linewidth}
                \centering                \includegraphics[width=\textwidth]{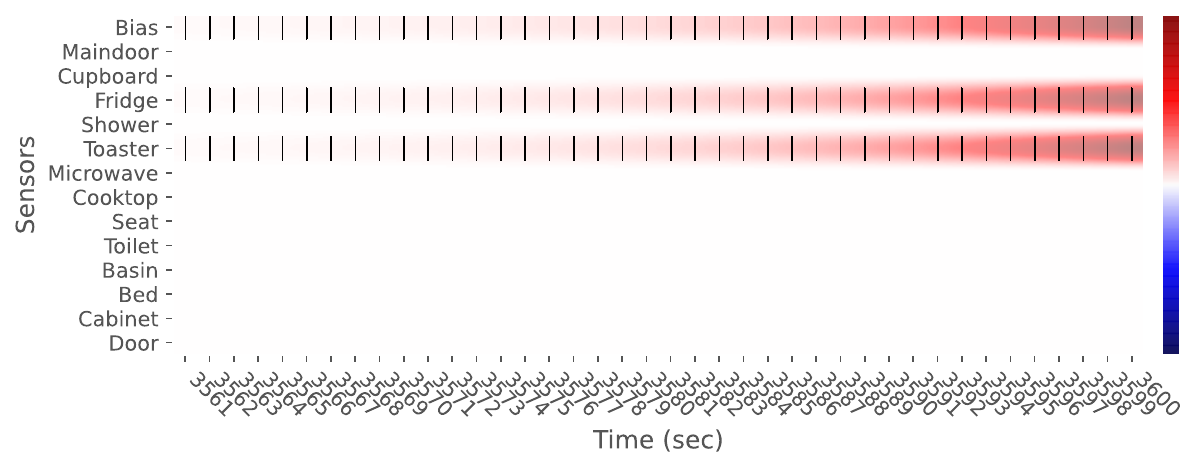}
                \caption{SAM breakfast class activation.}
                \label{fig:v_one_sam}
            \end{subfigure}
            \caption{Attribution maps of our method compared to SAM~\cite{kim2021} for the same example of a correct prediction of SNN-1L ($y=$Breakfast). \textit{Best viewed in color.}}
            \label{fig:example_expl_correct}
            \vspace{-1em}
        \end{figure}

Example visualizations for correct predictions of SNN-1L in the synthetic task are shown in Figure~\ref{fig:synthetic_viz}. Examples of the ADL task are shown in Figure~\ref{fig:example_expl_correct}.

        Overall, the explanations generated by TSA\textsuperscript{NS}, TSA\textsuperscript{S}, and SAM for SNN-1L seem quite similar: Recent time steps attribute stronger than time steps further in the past. All explanations also recognize the spiking input to be the important part, while TSA\textsuperscript{NS} also assigns attribution to non-spiking parts of the data. Both TSA\textsuperscript{S} and TSA\textsuperscript{NS} show different attribution strengths between different input dimensions for the same time step, while SAM seems to assign similar non-zero values to the same time steps regardless of the input dimension. Explanations generated by SAM do not differentiate between positive and negative attributions.
        
        \subsubsection{Misclassifications}
\begin{figure}[t]
            \centering
            \begin{subfigure}[b]{0.95\linewidth}
              \includegraphics[width=\textwidth]{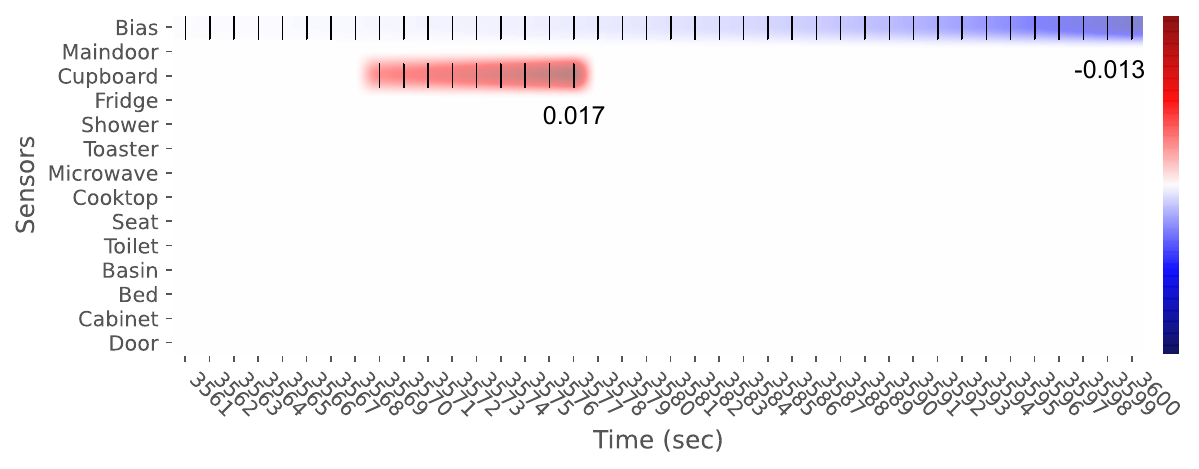}
                \caption{TSA\textsuperscript{S} lunch class activation.}
                \label{fig:v_one_s_incorrect_lunch}
            \end{subfigure}
            \hfill
            \begin{subfigure}[b]{0.95\linewidth}
                \includegraphics[width=\textwidth]{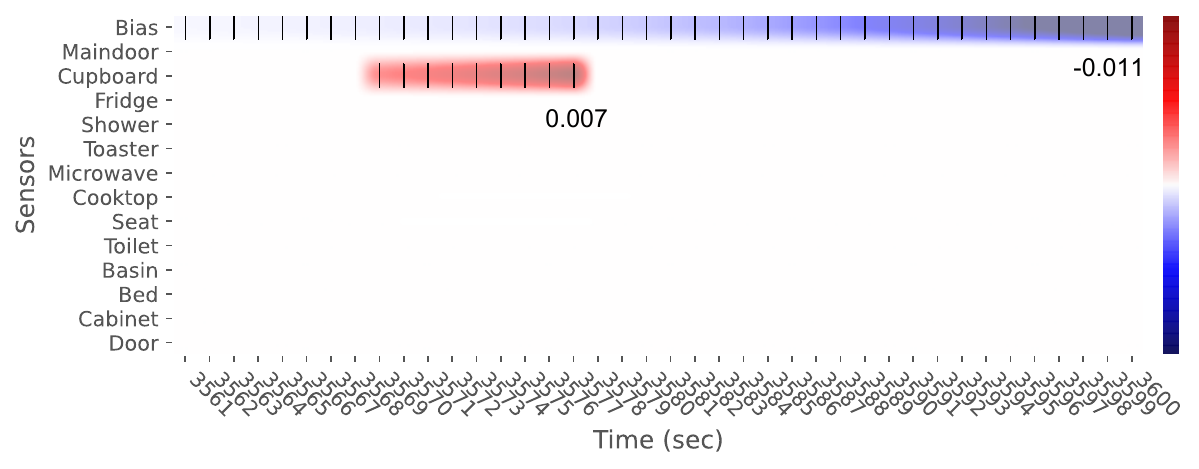}
                \caption{TSA\textsuperscript{S} breakfast class activation.}
                \label{fig:v_one_s_incorrect_breakfast}
            \end{subfigure}  
            \caption{Visualizations of TSA\textsuperscript{S} for an incorrect prediction. The numerical values at time steps 3577 and 3600 were added manually.´The attribution of the cupboard is stronger for the lunch class than the breakfast class. \textit{Best viewed in color.}}
            \label{fig: example_expl_incorrect}
            \vspace{-1em}
        \end{figure}

        In Figure~\ref{fig: example_expl_incorrect}, explanations for an incorrect prediction of the model are displayed for the true class (\textit{Breakfast}) as well as the predicted class (\textit{Lunch}). While both maps look very similar, there are subtle differences. In both cases, the \textit{Cupboard} sensor activation attributes positively to the classes. This makes sense because a kitchen sensor is likely to be connected to meal-related classes. Still, the map shows that the model connects this sensor's activation with the \textit{Lunch} class rather than the \textit{Breakfast} class at this point in the data. It is also noticeable that the model bias is negatively attributing to the prediction in both cases, with the negative attribution in the last time step being larger for \textit{Lunch}. However, in the \textit{Breakfast} class case, the (negative) bias is slightly stronger across time, suggesting that the model is biased against \textit{Breakfast}. Given the stronger positive attribution of the \textit{Cupboard} sensor activation and the slight bias against \textit{Breakfast}, \textit{Lunch}, a likely similar-looking class, is predicted. This example highlights the informativeness of negative and positive class attribution, as SAM would be unable to distinguish this.
        
        \subsubsection{Deep SNNs}
\begin{figure}[t]
            \centering
            \begin{subfigure}[b]{0.95\linewidth}
                \centering                \includegraphics[width=\textwidth]{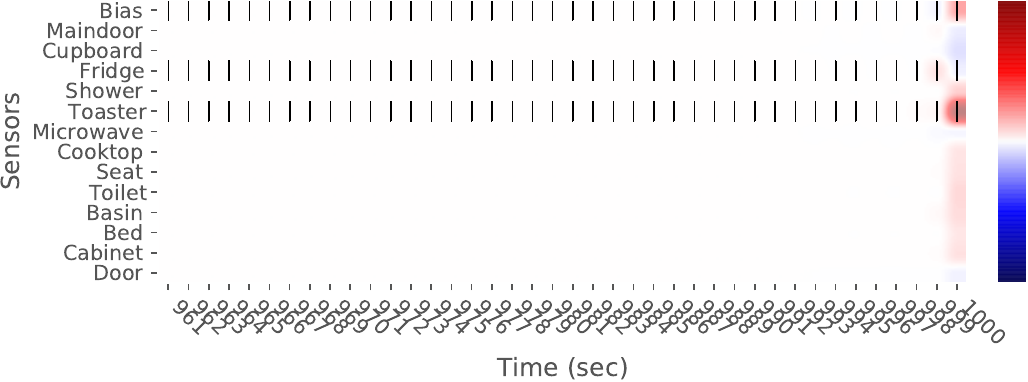}
                \caption{TSA\textsuperscript{NS} breakfast class activation for SNN-2L.}
                \label{fig:v_two_ns}        
            \end{subfigure}
            \hfill
            \begin{subfigure}[b]{0.95\linewidth}
                \centering                \includegraphics[width=\textwidth]{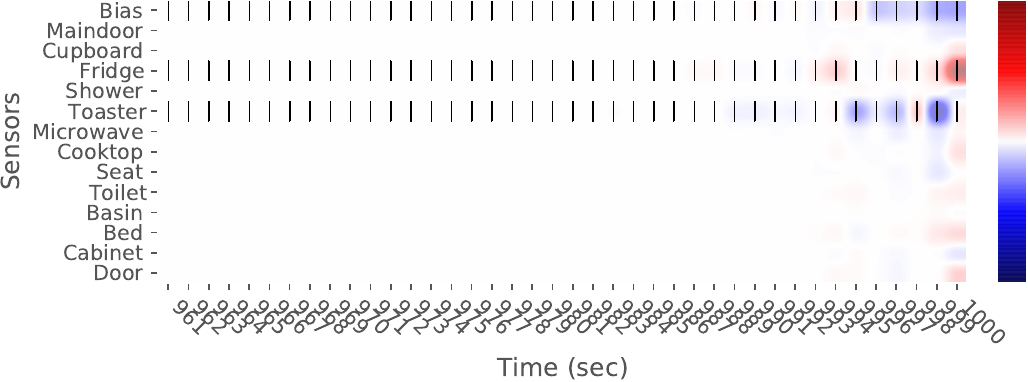}
                \caption{TSA\textsuperscript{NS} breakfast class activation for SNN-3L.}
                \label{fig:v_three_ns}            
            \end{subfigure}
            \caption{TSA\textsuperscript{NS} explanations from SNN-2L and SNN-3L on the same example of a correct prediction. \textit{Best viewed in color.}}
            \label{fig:example_expl_deepmodels}
            \vspace{-1em}
        \end{figure}

        Figure~\ref{fig:example_expl_deepmodels} shows examples from TSA\textsuperscript{NS} explanations extracted from SNN-2L and SNN-3L on the same sample as Figure~\ref{fig:example_expl_correct}. 
        The examples show that the decay rate $\gamma$ is important for SNN models, as it dictates how far into the past spikes can influence the model prediction at time $t$. SNN-2L and SNN-3L have different $\gamma$. SNN-2L, with a steeper decay, can only consider the last two to three time steps while SNN-3L can look further into the past. 
        
        Furthermore, the examples also show a limitation of TSA for deep models. Attribution values tend to alternate in an input dimension between positive and negative values, which could be a result of the repeated multiplication of values in $[-1, 1]$ during the aggregation of attribution scores across layers. This indicates a need to further study TSA on deep SNNs. 
    
\section{Discussion}
    \label{sec:discussion}
    The quantitative evaluation shows that TSA achieves a significant improvement over SAM in both the synthetic and ADL data sets. More specifically, TSA\textsuperscript{NS} outperforms SAM in all tested properties while TSA\textsuperscript{S} is better at explaining SNN-1L and roughly on par with SAM for explaining SNN-2L and SNN-3L. Overall,  no local explanation method outperforms all others in all tested properties, which demonstrates the multi-dimensionality of explanation quality. 
    
    In comparison to SAM, TSA\textsuperscript{S} and TSA\textsuperscript{NS} both consider the \textbf{model weights} in the computation of attribution directly. Since the weights are an essential part of a neural network, we hypothesized that their direct consideration would improve explanation quality. Our results show that this is largely the case, especially for explanations of SNN-1L. For this model, both TSA\textsuperscript{S} and TSA\textsuperscript{NS} showed significant improvements to SAM in all tested properties except for output-completeness in the synthetic task. 
    The continuity of TSA could also be positively impacted by considering weights $W$, as $W$ is a constant factor across time, which scales the NCS. 
    Our SNN models exhibit weights $W$ with weight values $|<1|$, hence the attribution values are kept small, leading to more compact explanations. In cases where weight values are large, weights could be normalized for explanation compactness. 
    Additionally, TSA distinguishes between positive and negative attribution because we consider the excitatory and inhibitory nature of $W$, whereas SAM is unable to make this distinction. This can also be observed in the examples shown (Figure~\ref{fig:v_one_sam}), where TSA\textsuperscript{S} and TSA\textsuperscript{NS} both assign a negative bias attribution while SAM marks the bias as positively attributing.  
    Nevertheless, $W$ could potentially cause the decreased explanation performance for TSA explanations of deeper SNNs. Signs may cancel each other out as weighted NCS are multiplied across layers.
    Furthermore, $W$ could lead to vanishing attribution for deep models in our case (e.g., indicated by the compactness of deep SNNs), due to repeated multiplication of values $|<1|$. This can be investigated in future work.
    
    The results show that \textbf{absent spikes are relevant}. We extended the definition of the NCS by the contribution score for absent spikes in TSA\textsuperscript{NS}. 
    The quantitative results clearly show that inactive input dimensions are relevant to the model prediction. While the models mainly use spikes for their prediction, the inactivity of other neurons is also relevant. Our synthetic data set with logical OR is designed such that the absence of spikes is decisive for the class label, and such reasoning should be correctly reflected in the explanation. In the real-world ADL data set, the active \textit{Toaster} sensor is for example important to the model at the same time as the bathroom sensors' inactivity to classify \textit{Breakfast}. As both correctness and output-completeness of TSA\textsuperscript{NS} present significant improvements to SAM and TSA\textsuperscript{S}, it is likely that the model has learned the connection between non-spikes and classes. Particularly the experiments with synthetic data verify that this observation is not specific to the models trained on the ADL task, but also holds for accurate temporally coded SNN models. 
    Since SAM and TSA\textsuperscript{S} do not consider absent spikes, they generally perform worse in the evaluation for content-related measures.
    The evaluation of continuity and compactness uses absolute values to determine the explanation's performance. As SAM and TSA\textsuperscript{S} are restricted to defining attribution only for spikes, the amount of change in the explanation is also limited. Therefore, TSA\textsuperscript{S} explanations are more compact than TSA\textsuperscript{NS}. However, both do not exhibit a large difference in continuity, which indicates that TSA itself is continuous. 
    
    \subsection{Limitations} 
    While the evaluation results indicate that TSA improves upon SAM in terms of explanation quality, there are limitations to consider. 
    First, TSA, like SAM, is a post-hoc explanation method. This means that it is applied to trained SNN models. Any unreasonable-looking or unexpected explanation could therefore either be rooted in erroneous model behavior or in errors of the explanation method~\cite{gilpin_explaining_2018}. 
    To ensure that the latter is not the case, we systematically and quantitatively evaluated TSA in our work, where we found that the applicability of TSA to deep SNNs likely requires further research. 
    Second, we tested only SNNs based on LIF. While TSA only relies on spike times, membrane potentials and weights and is therefore applicable to any spike generation mechanism, the computation of the NCS requires a specified decay parameter $\gamma$. With LIF and other integrate-and-fire models, the choice of $\gamma$ is straightforward. For more complex models that do not specify $\gamma$ directly (e.g. Hodgkin-Huxley neurons~\cite{HODGKIN.1952}), it must be defined first before using TSA to explain.
    Third, the evaluation does not consider user-related tasks since the technical feasibility of the method was in focus first. We did not explicitly test for comprehensibility, but it is an important property of explanations~\cite{Guidotti.2019}. Hence, TSA as an explanation method for SNNs does not have sufficient maturity for usage yet. Instead, it offers an effective starting point for further research on explanations for SNNs.

\section{Conclusion}
    \label{sec:conclusion}
    We present a local explanation method to address the outcome explanation problem for SNNs. We define \textit{Temporal Spike Attribution} (TSA), a feature attribution explanation that uses model-internal variables to explain the prediction of time series classification. 
    The two variants of TSA differ in the consideration of absent spike contribution, namely TSA\textsuperscript{S} and TSA\textsuperscript{NS}. We demonstrate TSA with three SNNs of different depths with temporal data and evaluated it in a multi-faceted quantitative analysis.
    We found that TSA\textsuperscript{NS} is superior in correctness and output-completeness, which shows the importance of considering absent spikes. There is no substantial difference between the TSA variants in terms of continuity and compactness. Besides, a decrease in the quality of TSA is observed for deep SNNs, which we attribute to the aggregation across layers. 
    We find that it is advantageous to consider all available information for explaining SNN predictions. Future studies could focus on the explanation of deep SNNs and human-comprehensibility based on TSA. 

\bibliographystyle{ieeetr}
\bibliography{bibliography}

\begin{thebibliography}{10}

\bibitem{Maass.1997}
W.~Maass, ``Networks of spiking neurons: The third generation of neural network models,'' {\em Neural Networks}, vol.~10, no.~9, pp.~1659--1671, 1997.

\bibitem{Gerstner.2014}
W.~Gerstner, W.~M. Kistler, R.~Naud, and L.~Paninski, {\em Neuronal dynamics: From single neurons to networks and models of cognition / Wulfram Gerstner, Werner M. Kistler, Richard Naud, Liam Paninski}.
\newblock Cambridge: {Cambridge University Press}, 2014.

\bibitem{Ponulak2011-tt}
F.~Ponulak and A.~Kasinski, ``Introduction to spiking neural networks: Information processing, learning and applications,'' {\em Acta Neurobiol. Exp. (Wars.)}, vol.~71, no.~4, pp.~409--433, 2011.

\bibitem{Wang.2020}
X.~Wang, X.~Lin, and X.~Dang, ``Supervised learning in spiking neural networks: A review of algorithms and evaluations,'' {\em Neural networks : the official journal of the International Neural Network Society}, vol.~125, pp.~258--280, 2020.

\bibitem{Murray.1998}
A.~F. Murray, {\em Pulse-Based Computation in VLSI Neural Networks}, pp.~87--–109.
\newblock Cambridge, MA, USA: MIT Press, 1999.

\bibitem{sharmin.2019}
S.~Sharmin, P.~Panda, S.~S. Sarwar, C.~Lee, W.~Ponghiran, and K.~Roy, ``A comprehensive analysis on adversarial robustness of spiking neural networks,'' in {\em 2019 International Joint Conference on Neural Networks (IJCNN)}, pp.~1--8, 2019.

\bibitem{Bing.2018}
Z.~Bing, C.~Meschede, F.~Röhrbein, K.~Huang, and A.~C. Knoll, ``A survey of robotics control based on learning-inspired spiking neural networks,'' {\em Frontiers in Neurorobotics}, vol.~12, 2018.

\bibitem{azghadi.2020}
M.~R. Azghadi, C.~Lammie, J.~K. Eshraghian, M.~Payvand, E.~Donati, B.~Linares-Barranco, and G.~Indiveri, ``Hardware implementation of deep network accelerators towards healthcare and biomedical applications,'' {\em IEEE Transactions on Biomedical Circuits and Systems}, vol.~14, no.~6, pp.~1138--1159, 2020.

\bibitem{He.2019}
J.~He, S.~L. Baxter, J.~Xu, J.~Xu, X.~Zhou, and K.~Zhang, ``The practical implementation of artificial intelligence technologies in medicine,'' {\em Nature medicine}, vol.~25, no.~1, pp.~30--36, 2019.

\bibitem{Adadi.2018}
A.~Adadi and M.~Berrada, ``{Peeking Inside the Black-Box: A Survey on Explainable Artificial Intelligence (XAI)},'' {\em IEEE Access}, vol.~6, pp.~52138--52160, 2018.

\bibitem{Guidotti.2019}
R.~Guidotti, A.~Monreale, S.~Ruggieri, F.~Turini, F.~Giannotti, and D.~Pedreschi, ``A survey of methods for explaining black box models,'' {\em ACM Computing Surveys}, vol.~51, no.~5, pp.~1--42, 2019.

\bibitem{Molnar.19102020}
C.~Molnar, G.~Casalicchio, and B.~Bischl, ``Interpretable machine learning -- a brief history, state-of-the-art and challenges,'' in {\em ECML PKDD 2020 Workshops}, (Cham), pp.~417--431, Springer International Publishing, 2020.

\bibitem{Jeyasothy.28022019}
A.~Jeyasothy, S.~Sundaram, S.~Ramasamy, and N.~Sundararajan, ``A novel method for extracting interpretable knowledge from a spiking neural classifier with time-varying synaptic weights,'' {\em CoRR}, vol.~abs/1904.11367, 2019.

\bibitem{kim2021}
Y.~Kim and P.~Panda, ``Visual explanations from spiking neural networks using inter-spike intervals,'' {\em Scientific Reports}, vol.~11, no.~1, 2021.

\bibitem{nauta2022anecdotal}
M.~Nauta, J.~Trienes, S.~Pathak, E.~Nguyen, M.~Peters, Y.~Schmitt, J.~Schl\"{o}tterer, M.~van Keulen, and C.~Seifert, ``From anecdotal evidence to quantitative evaluation methods: A systematic review on evaluating explainable ai,'' {\em ACM Comput. Surv.}, feb 2023.
\newblock Just Accepted.

\bibitem{pmlr-v80-kim18d}
B.~Kim, M.~Wattenberg, J.~Gilmer, C.~Cai, J.~Wexler, F.~Viegas, and R.~sayres, ``Interpretability beyond feature attribution: Quantitative testing with concept activation vectors ({TCAV}),'' in {\em Proceedings of the 35th International Conference on Machine Learning} (J.~Dy and A.~Krause, eds.), vol.~80 of {\em Proceedings of Machine Learning Research}, pp.~2668--2677, PMLR, 10--15 Jul 2018.

\bibitem{Nauta2021NeuralPT}
M.~Nauta, R.~van Bree, and C.~Seifert, ``Neural prototype trees for interpretable fine-grained image recognition,'' {\em 2021 IEEE/CVF Conference on Computer Vision and Pattern Recognition (CVPR)}, pp.~14928--14938, 2021.

\bibitem{ribeiro_2016_lime}
M.~T. Ribeiro, S.~Singh, and C.~Guestrin, ``"why should i trust you?": Explaining the predictions of any classifier,'' in {\em Proceedings of the 22nd ACM SIGKDD International Conference on Knowledge Discovery and Data Mining}, KDD '16, (New York, NY, USA), p.~1135–1144, Association for Computing Machinery, 2016.

\bibitem{lundberg_lee_2017_shap}
S.~M. Lundberg and S.-I. Lee, ``A unified approach to interpreting model predictions,'' in {\em Proceedings of the 31st International Conference on Neural Information Processing Systems}, NIPS'17, (Red Hook, NY, USA), p.~4768–4777, Curran Associates Inc., 2017.

\bibitem{Poyiadzi2021OnTO}
R.~Poyiadzi, X.~Renard, T.~Laugel, R.~Santos-Rodr{\'i}guez, and M.~Detyniecki, ``On the overlooked issue of defining explanation objectives for local-surrogate explainers,'' {\em ArXiv}, vol.~abs/2106.05810, 2021.

\bibitem{ijcai/BhattWM20}
U.~Bhatt, A.~Weller, and J.~M.~F. Moura, ``Evaluating and aggregating feature-based model explanations,'' in {\em Proceedings of the Twenty-Ninth International Joint Conference on Artificial Intelligence, {IJCAI} 2020} (C.~Bessiere, ed.), pp.~3016--3022, ijcai.org, 2020.

\bibitem{ijcai/0002C20}
R.~Srinivasan and A.~Chander, ``Explanation perspectives from the cognitive sciences - {A} survey,'' in {\em Proceedings of the Twenty-Ninth International Joint Conference on Artificial Intelligence, {IJCAI} 2020} (C.~Bessiere, ed.), pp.~4812--4818, ijcai.org, 2020.

\bibitem{liu_synthetic_2021}
Y.~Liu, S.~Khandagale, S.~Khandagale, C.~White, and W.~Neiswanger, ``Synthetic {Benchmarks} for {Scientific} {Research} in {Explainable} {Machine} {Learning},'' in {\em Proceedings of the {Neural} {Information} {Processing} {Systems} {Track} on {Datasets} and {Benchmarks}} (J.~Vanschoren and S.~Yeung, eds.), vol.~1, 2021.

\bibitem{Neftci.2019}
E.~O. Neftci, H.~Mostafa, and F.~Zenke, ``Surrogate gradient learning in spiking neural networks: Bringing the power of gradient-based optimization to spiking neural networks,'' {\em IEEE Signal Processing Magazine}, vol.~36, no.~6, pp.~51--63, 2019.

\bibitem{Ordonez.2013}
F.~J. Ord{\'o}{\~n}ez, P.~de~Toledo, and A.~Sanchis, ``Activity recognition using hybrid generative/discriminative models on home environments using binary sensors,'' {\em Sensors (Basel, Switzerland)}, vol.~13, no.~5, pp.~5460--5477, 2013.

\bibitem{Mosley.2013}
L.~Mosley, {\em A balanced approach to the multi-class imbalance problem}.
\newblock 2013.

\bibitem{Hamad.2021}
R.~A. Hamad, M.~Kimura, L.~Yang, W.~L. Woo, and B.~Wei, ``Dilated causal convolution with multi-head self attention for sensor human activity recognition,'' {\em Neural Computing and Applications}, 2021.

\bibitem{Montavon.2018}
G.~Montavon, W.~Samek, and K.~M{\"u}ller, ``Methods for interpreting and understanding deep neural networks,'' {\em Digit. Signal Process.}, vol.~73, pp.~1--15, 2018.

\bibitem{Schlegel.2019}
U.~Schlegel, H.~Arnout, M.~El-Assady, D.~Oelke, and D.~A. Keim, ``Towards a rigorous evaluation of xai methods on time series,'' in {\em 2019 IEEE/CVF International Conference on Computer Vision Workshop (ICCVW)}, pp.~4197--4201, 2019.

\bibitem{breiman2001random}
L.~Breiman, ``Random forests,'' {\em Machine learning}, vol.~45, no.~1, pp.~5--32, 2001.

\bibitem{yeh.2019}
C.-K. Yeh, C.-Y. Hsieh, A.~Suggala, D.~I. Inouye, and P.~K. Ravikumar, ``On the (in)fidelity and sensitivity of explanations,'' in {\em Advances in Neural Information Processing Systems} (H.~Wallach, H.~Larochelle, A.~Beygelzimer, F.~d\textquotesingle Alch\'{e}-Buc, E.~Fox, and R.~Garnett, eds.), vol.~32, Curran Associates, Inc., 2019.

\bibitem{gilpin_explaining_2018}
L.~H. Gilpin, D.~Bau, B.~Z. Yuan, A.~Bajwa, M.~Specter, and L.~Kagal, ``Explaining explanations: An overview of interpretability of machine learning,'' in {\em 2018 IEEE 5th International Conference on Data Science and Advanced Analytics (DSAA)}, pp.~80--89, 2018.

\bibitem{HODGKIN.1952}
A.~L. Hodgkin and A.~F. Huxley, ``A quantitative description of membrane current and its application to conduction and excitation in nerve,'' {\em The Journal of physiology}, vol.~117, no.~4, pp.~500--544, 1952.

\bibitem{Zenke.2018}
F.~Zenke and S.~Ganguli, ``Superspike: Supervised learning in multilayer spiking neural networks,'' {\em Neural computation}, vol.~30, no.~6, pp.~1514--1541, 2018.

\end{thebibliography}

 \appendix

    \section{Details on Data Sets}
     All code for synthetic data set generation as well as preprocessing of real-world data is available in our Github repository \url{https://github.com/ElisaNguyen/tsa-explanations}.
     \label{appendix:data}
    
         \subsection{Synthetic Data}
         \label{appendix:data:synthetic}
         The synthetic data set is generated as a smaller and simpler version of the real-world data, which shall be easy to learn. It consists of two-dimensional time series data with four classes. The data is binary (i.e., $x_{i,t} \in \{0,1\}$). We generate the data by sampling both the duration and activation of $x_i$ at random. The maximum duration for an activity is set at 600 time steps, and we generate 900 000 time steps sequentially as the entire data set. Once the data is generated, we add labels per time step according to the data, as specified in Figure~\ref{fig:syn_data} of the main paper. There are no misclassified time steps. 
        
         \subsection{Real-world Data}
         \label{appendix:data:real}
         The real-world data set we use is the Binary ADL data set~\cite{Ordonez.2013}, which is openly available in the UCI machine learning repository. The data set consists of the start and end times of activities and their corresponding labels across two subjects, A and B. Given this information, we generate continuous time series across the complete duration of the data set. Gaps between activities were filled with inactivity in all sensors. We introduce the \textit{Other} class for these time gaps.
         We found two cases where the activity end precedes the start (Index 78, 80 of subject A). We excluded these activities from the data set as either the data collection or labeling is faulty and filled their time with inactivity.

     \section{SNN Model Definition and Training}
         \label{appendix:snn_def_tuning_training}
         The SNN models (\textit{SNN-1L}, \textit{SNN-2L} and \textit{SNN-3L}) were built as fully connected recurrent networks with binary activations, using discretized formulas of the network dynamics, in accordance with \cite{Neftci.2019}. Likewise, the models are also trained with surrogate gradient descent using a fast sigmoid surrogate. Moreover, we adapt the training procedure to our data set that exhibits a temporal order (i.e., activities follow one after another in time). The membrane potential state $u(t)$ is retained between data samples to reflect the temporal dependencies. So, state variables of the model are initialized with the last states of the last simulation. 
         The maximum membrane potential at the output layer at each time step $\Delta t$ determines the prediction. This allows the use of regular loss functions for optimization. Similar to \cite{Zenke.2018}, the models are optimized with negative log-likelihood.
         While the main focus of our work is explaining SNNs and not their optimization, the models should demonstrate a clear improvement in predictive performance to pure chance and perform reasonably well, so that the models have learned something worth explaining. 
        
         Due to the simplicity of the synthetic classification task, we omit hyperparameter tuning. The model hyperparameters were determined across all models beforehand (Table~\ref{tab:synparams})
         \begin{table}[t]
                 \centering
                     \begin{tabular}{@{}ll@{}}
                     \toprule
                     \textbf{Hyperparameter} & \textbf{Choice} \\ \midrule
                     $\Delta t$ & 0.001 \\
                     $\tau_{syn}$                & 0.01           \\
                    $\tau_{mem}$                & 0.001                \\
                     Learning rate          & 0.001                       \\
                     Batch size             & 128            \\
                     Size of hidden layer &   10 \\ 
                     \bottomrule
                     \end{tabular}
                     \caption{Hyperparameters used for model building with synthetic data set.}
                 \label{tab:synparams}
             \end{table}

         Standard hyperparameters are not sufficient in the ADL task. Therefore, the hyperparameters of the networks are tuned in a greedy optimization process for 20 epochs under the assumption of independence for this task.
         With the tuned hyperparameters, the final models were fully retrained on the training set. As regularization, early stopping with a patience of 10 epochs was used, monitoring the validation loss. The final hyperparameters are shown in Table~\ref{tab:tuning}.
             \begin{table}[ht]
                 \centering
                 \begin{small}
                     \begin{tabular}{@{}llll@{}}
                     \toprule
                     \textbf{Hyperparameter} & \textbf{SNN-1L} & \textbf{SNN-2L} & \textbf{SNN-3L} \\ \midrule
                     $\Delta t$ & 0.001 & 0.001 & 0.001 \\
                     $\tau_{syn}$                & 0.01           & 0.01         & 0.01      \\
                     $\tau_{mem}$                & 0.01          & 0.001          & 0.01           \\
                     Learning rate          & 0.01           & 0.001          & 0.001             \\
                     Batch size             & 128        & 256         & 512           \\
                     Size of hidden layer 1 & -           &   100       & 50          \\
                     Size of hidden layer 2 & -           & -           & 25         \\ \bottomrule
                     \end{tabular}
                 \end{small}
                     \caption{Results of greedy hyperparameter optimization for all models on the binary ADL task. Tested ranges are: \{0.01, 0.001, 0.0001\} for $\Delta t$ and learning rate, \{0.1, 0.01, 0.001\} for $\tau_{syn}$ and $\tau_{mem}$, \{128, 256, 512\} for batch size, \{25, 50, 100, 200\} for hidden layer sizes.}
                 \label{tab:tuning}
             \end{table}
            
         The scripts used for hyperparameter optimization and model training are provided in our GitHub repository \url{https://github.com/ElisaNguyen/tsa-explanations}. 
            
     \section{Evaluation Measure Computation}
             \label{appendix:metrics}
             In this section, pseudocode and formulas are given for this work's quantitative evaluation of explainability performance. 
             The scripts used in our experiments are also published in our GitHub repository (\texttt{https://github.com/ElisaNguyen/tsa-explanations}).
            
             \textbf{Correctness} is computed in explanation selectivity~\cite{Montavon.2018}, shown in algorithm~\ref{alg:explanation_selectivity}. The deletion of feature segments is performed as an inversion of the time series, in line with~\cite{Schlegel.2019}.
             \begin{algorithm}
                     \caption{\textbf{Explanation selectivity}}\label{alg:explanation_selectivity}
                     \begin{algorithmic}
                         \STATE Let $e$ be the explanation function that results in feature attribution map $A(x, t)$ describing the attributions to the predicted class, $f(x, t)$ denote the model's prediction on an input $x\in X$ at time $t$. Let $R$ be the total number of feature segments of $x$, $N$ the size of the test set $X$ and $Y$ be the corresponding ground-truth labels for $X$. 
                        
                         \FOR{$x\in X$}
                             \FOR{$t = 1,2,...,T$ with $T$ being the duration of $x$}
                                 \STATE $A(x, t) \gets e(f, x, t)$
                                 \STATE Define $R$ feature segments.
                                 \STATE Sort the feature segments in descending order by their mean attribution values.
                                
                                 \FOR{rank $r=0,1,...,R$}
                                     \STATE $x^{\text{inv@}r} \gets$ Invert feature segment $x^{(r)}$ so that $x^{(r)}=|x^{(r)}-1|$.
                                     \STATE $\hat{y}^{\text{inv@}r}\gets f(x^{\text{inv@}r}, t)$
                                 \ENDFOR
                             \ENDFOR
                         \ENDFOR
                         \STATE Let $X^{\text{inv@}r}$ denote $X$ with feature segments up to rank $r$ inverted.
                         \FOR{rank $r=0,1,...,R$}
                             \STATE Compute Balanced Accuracy of $\hat{Y}^{\text{inv@}r}, Y$. 
                         \ENDFOR
                         \STATE Compute explanation selectivity as the AUC of the graph resulting from the performance of the model depending on the amount of feature segments inverted. 
                     \end{algorithmic}
                 \end{algorithm}
                
         \textbf{Output-completeness} is measured with a model preservation check upon deletion of unimportant feature segments, specified in algorithm~\ref{alg:completeness}. 
        
             \begin{algorithm}
                     \caption{\textbf{Model preservation check}}\label{alg:completeness}
                     \begin{algorithmic}
                         \STATE Let $f(x, t)$ be a SNN model's prediction on input $x\in X$ at time $t$, $e$ be the explanation function which results in attribution map $A(x, t)$ that describes the attribution to the predicted class. Let $\epsilon$ be the threshold for feature importance (in our experiments $\epsilon = 0$). 
                         \FOR{$x\in X$}
                             \FOR{$t = 1,2,...,T$ with $T$ being the duration of $x$}
                                 \STATE $A(x, t) \gets e(f, x, t)$.
                                 \STATE Mask $A$ where $|a|>\epsilon$.
                                 \STATE $x_p\gets$ Perturb unmasked area of $A$. 
                                 \STATE $\hat{y}_p \gets f(x_p, t)$
                             \ENDFOR
                        \ENDFOR
                         \STATE Compute balanced accuracy of $\hat{Y}_p, \hat{Y}$ as a model preservation check. 
                     \end{algorithmic}
                 \end{algorithm} 
                
         \textbf{Continuity} is measured in max-sensitivity\cite{yeh.2019}, denoted as Sens$_{\textrm{max}}$:
         \begin{small}
             \begin{equation}
                 \label{eq: max-sensitivity}
                 \textrm{Sens}_{\textrm{max}}(e, f, x, t, r) = \max_{||x'-x|| \leq r}||e(f,x', t)-e(f,x, t)||
             \end{equation}
         \end{small}
        
         where $e$ refers to the explanation, $f$ to the model, $x$ the input data, $t$ the current timestep and $r$ defines the neighborhood region in which perturbed data $x'$ is still considered as similar to $x$. 
        
         \textbf{Compactness} is computed as follows:  
         \begin{equation}
             \textrm{Compactness}(A) = \frac{1}{N}\sum_i^D\sum_j^t |a_{i,j}|
         \end{equation}
         where $A$ is the matrix of dimension $D\times t$, i.e. input dimensionality $D \times$ the timestep to be explained. $N$ denotes the total number of explanations extracted for the experiment.

\end{document}